\documentclass[acmtog]{acmart}
\makeatletter
\renewcommand\@formatdoi[1]{\ignorespaces}
\makeatother

\usepackage{booktabs} 
\usepackage{multirow} 
\usepackage{multicol} 
\usepackage{amsmath}
\usepackage{hyperref}
\usepackage{pifont}
\newcommand{\cmark}{\ding{51}}%
\newcommand{\xmark}{\ding{55}}%

\usepackage{adjustbox}
\usepackage{array}

\newcolumntype{R}[2]{%
    >{\adjustbox{angle=#1,lap=\width-(#2)}\bgroup}%
    l%
    <{\egroup}%
}

\usepackage[ruled]{algorithm2e} 

\SetAlFnt{\small}
\SetAlCapFnt{\small}
\SetAlCapNameFnt{\small}
\SetAlCapHSkip{0pt}

\acmJournal{TOG}

\renewcommand\footnotetextcopyrightpermission[1]{} 
\begin{document}
\title{{\textsc acorn}: Adaptive Coordinate Networks for Neural Scene Representation}

\author{Julien N. P. Martel*, David B. Lindell*, Connor Z. Lin, Eric R. Chan, Marco Monteiro \lowercase{and} Gordon Wetzstein}
\affiliation{%
  \institution{Stanford University}
  \city{Stanford}
  \state{CA}
  \country{USA}
}
\email{jnmartel@stanford.edu}
\email{lindell@stanford.edu}
\thanks{*~J.~N.~P. Martel and D.~B. Lindell equally contributed. \\
Project webpage: http://computationalimaging.org/publications/acorn}

\renewcommand\shortauthors{Martel, J. N. P. and Lindell, D. B. et al.}
\newcommand{\jm}[1]{#1}
\newcommand{\highlight}[1]{#1}
\newcommand{\gw}[1]{#1}
\newcommand{\ec}[1]{#1}

\newcommand{\interp}{\textsc{LinInterp}}
\newcommand{\coarse}{\mathbf{x}_\text{g}}
\newcommand{\fine}{\mathbf{x}_\text{l}}
\newcommand{\encoder}{\Phi}
\newcommand{\decoder}{\Psi}
\newcommand{\out}{\bold{y}}
\newcommand{\gt}{\text{GT}}
\newcommand{\din}{d_{\text{in}}}
\newcommand{\dout}{d_{\text{out}}}
\newcommand{\blockerr}{E}

\newcommand{\imerge}{I^{\uparrow}}
\newcommand{\isplit}{I^{\downarrow}}
\newcommand{\inone}{I^{=}}
\newcommand{\igrp}{I^{g}}
\newcommand{\tp}{\intercal}
\newcommand{\partition}{\mathcal{B}}

\begin{abstract}
Neural representations have emerged as a new paradigm for applications in rendering, imaging, geometric modeling, and simulation.
Compared to traditional representations such as meshes, point clouds, or volumes they can be flexibly incorporated into differentiable learning-based pipelines.
While recent improvements to neural representations now make it possible to represent signals with fine details at moderate resolutions (e.g., for images and 3D shapes), adequately representing large-scale or complex scenes has proven a challenge.
Current neural representations fail to accurately represent images at resolutions greater than a megapixel or 3D scenes with more than a few hundred thousand polygons.
Here, we introduce a new \jm{hybrid implicit--explicit} network architecture and training strategy that adaptively allocates resources during training and inference based on the local complexity of a signal of interest.
Our approach uses a multiscale block-coordinate decomposition, similar to a quadtree or octree, that is optimized during training.
The network architecture operates in two stages: using the bulk of the network parameters, a coordinate encoder generates a feature grid in a single forward pass.
Then, hundreds or thousands of samples within each block can be efficiently evaluated using a lightweight feature decoder.
With this \jm{hybrid implicit--explicit} network architecture, we demonstrate the first experiments that fit gigapixel images to nearly 40 dB peak signal-to-noise ratio.
Notably this represents an increase in scale of over 1000$\times$ compared to the resolution of previously demonstrated image-fitting experiments.
Moreover, our approach is able to represent 3D shapes significantly faster and better than previous techniques; it reduces training times from days to hours or minutes and memory requirements by over an order of magnitude.
\end{abstract}


%


\begin{teaserfigure}
    \centering
    \includegraphics[width=\textwidth]{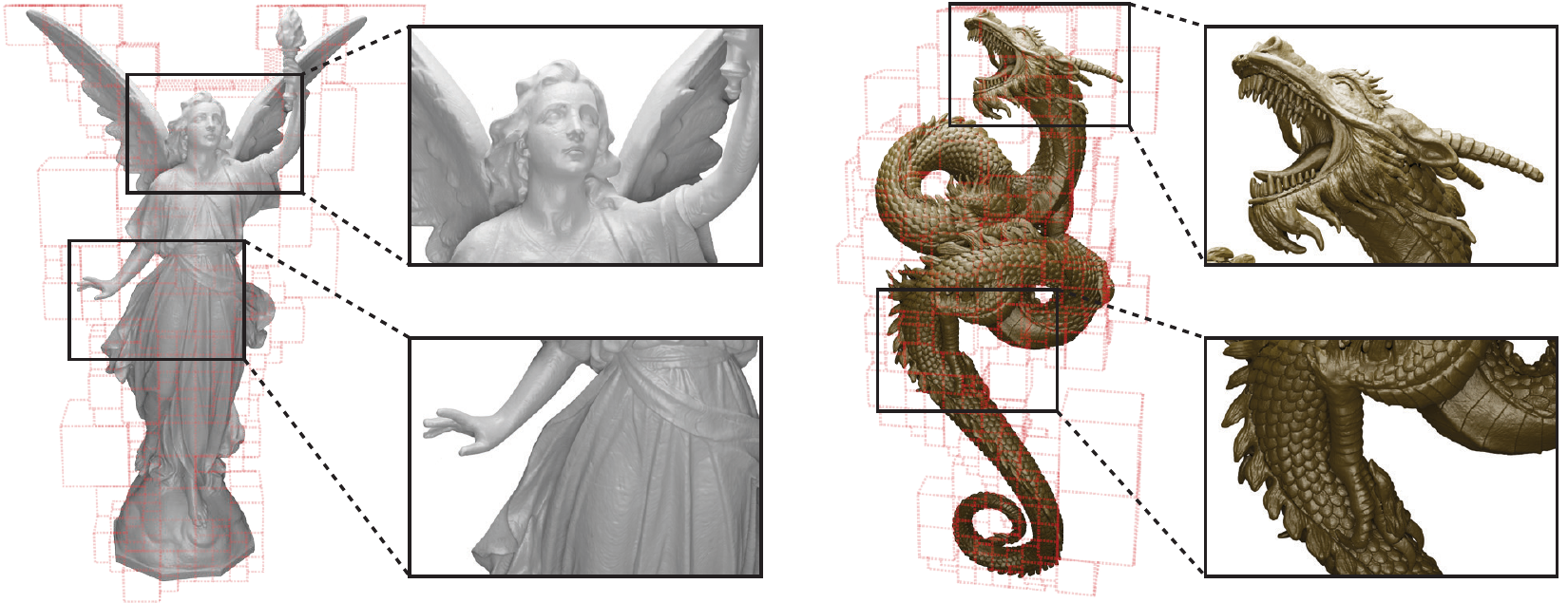}
    \caption{Adaptive coordinate networks for neural scene representation (\textsc{acorn}), can fit signals such as three-dimensional occupancy fields with high accuracy. Here we demonstrate fitting a 3D occupancy field where inputs to the model are continuous coordinates in space, and outputs are the occupancy at those positions. \textsc{acorn} optimizes a partition of space while training by allocating more blocks to regions with fine details. Shown here are two detailed 3D scenes obtained with our architecture, Lucy from the Stanford 3D Scanning Repository and a detailed model of a dragon as well as their optimized decompositions at the end of training.}
    \label{fig:teaser}
\end{teaserfigure}

\maketitle
\thispagestyle{empty}

\pagestyle{plain}

\section{Introduction}
\label{sec:introduction}
How we represent signals has a tremendous impact on how we solve problems in graphics, computer vision, and beyond. Traditional graphics representations, such as meshes, point clouds, or volumes, are well established as robust and general-purpose tools in a plethora of applications ranging from visual effects and computer games to architecture, computer-aided design, and 3D computer vision. Traditional representations, however, are not necessarily well-suited for emerging applications in neural rendering, imaging, geometric modeling, and simulation. For these emerging applications, we require representations that are end-to-end differentiable, quickly optimizable, and scalable in learning-based pipelines. Traditional representations are often at odds with these properties since they scale according to Nyquist sampling requirements and require storing spatio-temporal samples explicitly in memory. 

A variety of neural scene representations have been proposed over the last few years, which overcome some of these limitations (see Sec.~\ref{sec:related} for a detailed discussion). Among these, explicit neural representations usually focus on mimicking the functionality of traditional discrete representations in a differentiable manner, often using discrete grids of learned features. Implicit representations, or coordinate networks, parameterize images, shapes, or other signals using neural networks that take a spatio-temporal coordinate as input and output a quantity of interest, such as color, volume density, or a general feature. Coordinate networks are continuous, and the complexity of the signals they represent is limited by the capacity of the representation network rather than the grid size of an explicit representation.

While emerging neural scene representations are promising building blocks for a variety of applications, one of the fundamental challenges of these representations is their inability to scale to complex scenes. For example, although explicit representations are fast to evaluate, their memory requirements often scale unfavorably since they require storing a large number of explicit features. Coordinate networks can be more memory efficient, because they do not necessarily need to store features on a grid, but they are computationally inefficient; for every coordinate that is evaluated (e.g., every pixel of an image or every point in a 3D space), an entire feedforward pass through the underlying representation network has to be computed. Therefore, existing neural representations are either computationally inefficient or memory intensive, which prevents them from scaling to large-scale or complex scenes.

Here, we propose a new hybrid implicit--explicit coordinate network architecture for neural signal representations. Our architecture automatically subdivides the domain of the signal into blocks that are organized in a multiscale fashion. The signal is modeled by a coordinate network that takes a multiscale block coordinate as input and outputs a quantity of interest continuously within the block. Unlike existing local implicit representations~\jm{\cite{peng2020convolutional,chabra2020deep,jiang2020local,chen2020learning,kellnhofer2021lumigraph}}, our architecture does not map the input coordinate directly to the output, but to an intermediate low-resolution grid of local features using a \emph{coordinate encoder}. Using a separate but small \emph{feature decoder}, these features can be efficiently interpolated and decoded to evaluate the representation at any continuous coordinate within the block.
The benefit of this two-stage approach is that the bulk of the computation (generating feature grids) is performed only once per block by the coordinate encoder, and querying points within the block requires minimal computational overhead using the feature decoder.
This design choice makes our representation much faster to evaluate and more memory efficient than existing explicit, coordinate-based, or hybrid solutions.
\begin{table}[t!]
\begin{tabular}{lcccc}
& {Comp. Eff.} & {Mem. Eff.} & {Multiscale} & {Pruning}  \\ 
\toprule
Explicit 				& \cmark & \xmark & \xmark & \xmark \\ 
Global Implicit & \xmark & \cmark & \xmark & \xmark \\ 
Local Implicit 	& \cmark or \xmark & \xmark~or \cmark & \xmark & \xmark \\ 
NSVF 						& \cmark or \xmark & \xmark~or \cmark & \xmark & \cmark \\ 
Ours 						& \cmark & \cmark & \cmark & \cmark \\ 
\bottomrule
\end{tabular}
\caption{\label{tab:related}Existing explicit representations are memory inefficient and implicit coordinate networks are computationally inefficient, as they require a full forward pass for each coordinate they evaluate. Local implicit approaches and neural sparse voxel fields (NSVF)~\cite{liu2020neural} make a direct tradeoff between memory and computational efficiency, as they require an additional volume with latent codes or features to be stored. Our multiscale representation is computational and memory efficient, it allows us to prune empty space in an optimized manner, and we demonstrate its application to large-scale image and complex 3D shape representation.}
\vspace{-2.5em}
\end{table}
\begin{figure*}[t!]
    \includegraphics[]{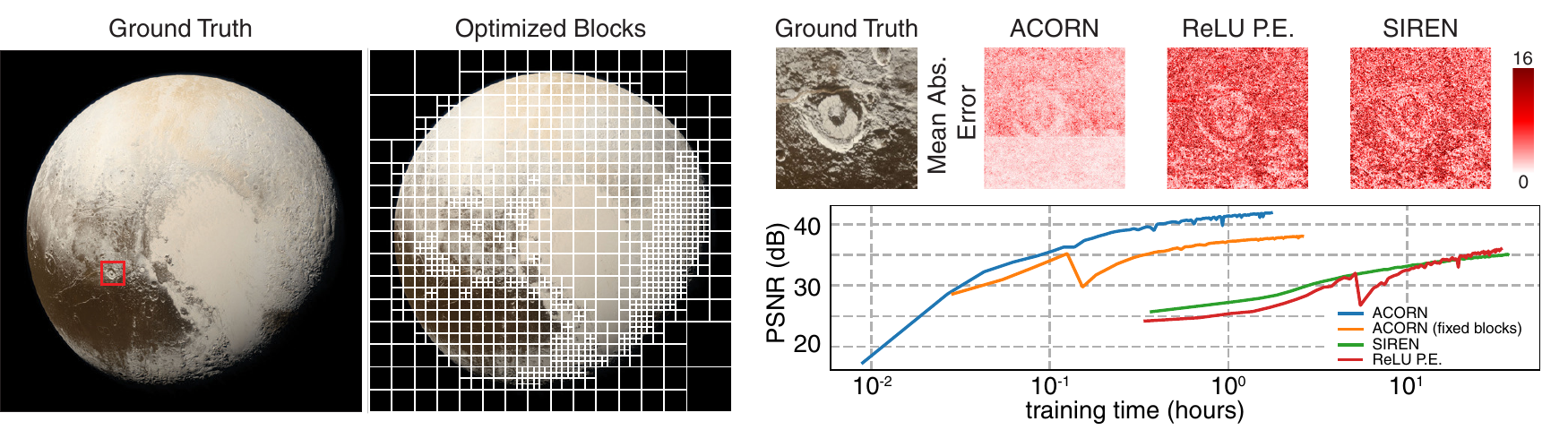}
    \caption{Large-scale image fitting. We fit networks to a 64 MP image of the dwarf planet Pluto, captured by the New Horizons space probe. The optimized block decomposition for our method is shown along with error maps for an inset of the image for our method, as well as a ReLU P.E. network~\cite{mildenhall2020nerf} and a SIREN~\cite{sitzmann2020siren}. All networks use roughly 9.5 M parameters. A plot of image PSNR vs. training time for all networks (including ours with a fixed, uniform block decomposition) is shown from 1,000 to 100,000 iterations. Our multiscale representation network converges to a PSNR of >40~dB in about one hour, whereas other representations struggle to adequately fit this image in more than one day.}
		\label{fig:pluto}
        \vspace{-1em}
\end{figure*}

Moreover, we propose a novel training strategy that adaptively refines the scale and locations of the block coordinates during training. 
The resulting block decomposition is similar in spirit to a quadtree or an \jm{octree} (Fig.~\ref{fig:pluto}, center left).
Determining the optimal block decomposition is a resource allocation problem that involves solving an integer linear program periodically during training.  
Notably, optimizing the block partitioning does not require pre-processing of the input signal and runs online using the training loss within each block. 
Effectively, this approach prunes empty space and allocates the capacity of the representation network in an optimal multiscale fashion. We also note that a single network is shared across all blocks of a scene, regardless of position or scale. 

The design choices discussed above make our representation both computationally efficient (i.e., by pruning empty space, a hierarchical subdivision, and using block-wise coordinate networks) and memory efficient (i.e., by the hybrid implicit--explicit network design). Our representation trains significantly faster than other coordinate networks and it does not spend resources on redundant parts of the signal. The inference time varies over the scene and depends on its local complexity. These benefits allow us to demonstrate the first experiments that fit gigapixel images as well as complex 3D scenes with a neural scene representation. For example, the largest image we processed (see Sec.~\ref{sec:images}) has a resolution of 51,200 $\times$ 19,456 pixels, which is about 1,000$\times$ larger than what has been demonstrated to date, and our approach is able to fit this image at a quality of nearly 40~dB peak signal-to-noise ratio (PSNR) within just a few hours. Similarly, we demonstrate that our approach is able to represent 3D shapes significantly faster and better than existing neural scene representations.

Our approach is a core building block of emerging neural approaches to imaging, geometric modeling, rendering, and simulation. We expect it to be of fundamental importance to many different areas in computer graphics and interactive techniques. \highlight{Source code is publicly available\footnote{\url{https://github.com/computational-imaging/ACORN}}.}

Specifically, our contributions include
\begin{itemize}
    \item a new multi-scale hybrid implicit--explicit signal representation network architecture,
		\item \jm{a training strategy that integrates} an integer linear program for automatic multiscale resource allocation and pruning,     
    \item state-of-the-art results for representing large-scale images and complex 3D scenes using neural representations.
\end{itemize}

\section{Related Work}
\label{sec:related}
A large body of work has been published in the area of neural scene representation. This section provides a best attempt to concisely overview this emerging research area.

\paragraph{Neural Scene Representations}

Among the recently proposed neural scene representations, \emph{explicit} representations directly represent 2D or 3D scenes using imperfect meshes~\cite{hedman2018deep,thies2019deferred,riegler2020free,zhang2020neural}, multi-plane~\cite{Zhou:2018,Mildenhall:2019,flynn2019deepview} or multi-sphere~\cite{Broxton:2020,Attal:2020:ECCV} images, or using a voxel grid of features~\cite{sitzmann2019deepvoxels,Lombardi:2019}. The primary benefit of explicit representations is that they can be quickly evaluated at any position and can thus be rendered fast. However, explicit grid-based representations typically come with a cost of large memory requirements and are thus challenging to scale. 

Alternatively, coordinate networks \emph{implicitly} define a 2D or 3D scene using neural networks. Many variants of these networks have been proposed in the last few years for modeling 3D shapes~\cite{atzmon2019sal,chabra2020deep,chen2019learning,davies2020overfit,genova2019deep,gropp2020implicit,park2019deepsdf,mescheder2019occupancy,michalkiewicz2019implicit,sitzmann2020siren}, view synthesis~\cite{chan2020pi,eslami2018neural,mildenhall2020nerf,Niemeyer2020CVPR,jiang2020sdfdiff,henzler2019platonicgan,nguyenphuoc2020blockgan,graf,sitzmann2019srns,yariv2020multiview,liu2020dist}, semantic labeling~\cite{kohli2020inferring}, and texture synthesis~\cite{Oechsle2019ICCV,saito2019pifu}.
The shared characteristic of all of these approaches is that signals or scenes are implicitly defined by a network that can be queried for continuous coordinates, rather than on a discrete pixel or voxel grid. This implies that the complexity of the scene is limited by the capacity of the network rather than the resolution of a grid. To represent a reasonably complex scene, however, coordinate networks must be fairly large, which makes them slow to query because an entire feedforward pass through the network needs to be computed for evaluating every single coordinate. This limitation makes current implicit representations difficult to scale.


\jm{The closest existing approaches to ours are local implicit methods. With applications to fitting shapes~\cite{peng2020convolutional,chabra2020deep,jiang2020local} or images~\cite{chen2020learning}, they use implicit networks to predict local or part-level scene information}. In these models, a single globally shared decoder network operates on local coordinates within some explicit grid representation. Each grid cell needs to store a latent code or feature vector that the local network is conditioned with. These methods largely differ in how this latent code vector is estimated, either \jm{through iterative optimization~\cite{chabra2020deep,jiang2020local}} or directly by a convolutional encoder network operating on a point cloud~\cite{peng2020convolutional}. In all cases, additional information, such as the volume of latent code vectors or the input point cloud, have to be stored along with the implicit network, which makes these methods less memory efficient and less scalable than a fully implicit representation that requires coordinate inputs alone. Similar to our approach, neural sparse voxel fields (NSVF)~\cite{liu2020neural} prunes empty space so that the representation network does not have to be evaluated in these regions. NSVF uses a heuristic coarse-to-fine sampling strategy to determine which regions are empty but, similar to local implicit models, requires storing an explicit volume of features representing the scene at a single scale. 

As outlined by Table~\ref{tab:related}, our network architecture is different from these approaches in several important ways.
First, our method is the only multiscale representation. This allows us to use the network capacity more efficiently and also to prune empty space in an optimized multiscale fashion.
Second, we realize that the computational bottleneck of all implicit networks is the requirement to compute many forward passes through the networks, one per coordinate that is evaluated.
This is also true for local implicit approaches. Instead, our architecture efficiently evaluates coordinates at the block level.
Third, our architecture is a ``pure'' coordinate network in that it does not require any additional features as input, so it is more memory efficient than approaches that store an explicit grid of features or latent codes.

\paragraph{Neural Graphics}

Neural rendering describes end-to-end differentiable approaches for projecting a 3D neural scene representation into one or multiple 2D images. Inverse approaches that aim at recovering a scene representation from 2D images require this capability to backpropagate the error of an image rendered from some representation with respect to a reference all the way into the representation to optimize it. Each type of representation may require a unique rendering approach that is most appropriate. For example, implicitly defined signed distance functions (SDFs) use ray marching or sphere tracing--based neural rendering (e.g.,~\cite{sitzmann2019srns,jiang2020sdfdiff,liu2020dist,yariv2020multiview}) whereas neural volume representations use a differentiable form of volume rendering \jm{(e.g.,~\cite{sitzmann2019deepvoxels,Lombardi:2019,mildenhall2020nerf,liu2020neural,niemeyer2020differentiable})}. A more detailed overview of recent approaches to neural rendering can be found in the survey by Tewari et al.~\shortcite{tewari2020state}. Similarly, neural representations and methods have recently been proposed for simulation problems~\cite{Kim:2019:deepfluids,li2020fourier,sitzmann2020siren}.

The multiscale neural representation we propose is agnostic to the type of signal it represents and we demonstrate that it applies to different signal types, including 2D images and 3D shapes represented as occupancy networks~\cite{mescheder2019occupancy}. Thus, our representation is complementary to and, in principle, compatible with most existing neural rendering and graphics approaches. Yet, a detailed exploration of all of these applications is beyond the scope of this paper. 



\paragraph{Multiscale Signal Representations}

Our multiscale network architecture is inspired by several common techniques used in other domains of graphics and interactive techniques, including adaptive mesh refinement used for fluid solvers or solving other partial differential equations (e.g.,~\cite{berger1984adaptive,huang2010adaptive}), bounding volume hierarchies used for collision detection or raytracing, and multiscale decompositions, such as wavelet transforms, that represent discrete signals efficiently at different scales.
%

\section{Multiscale Coordinate Networks}
\label{sec:methods}
\begin{figure*}[t!]
\includegraphics[width=\linewidth]{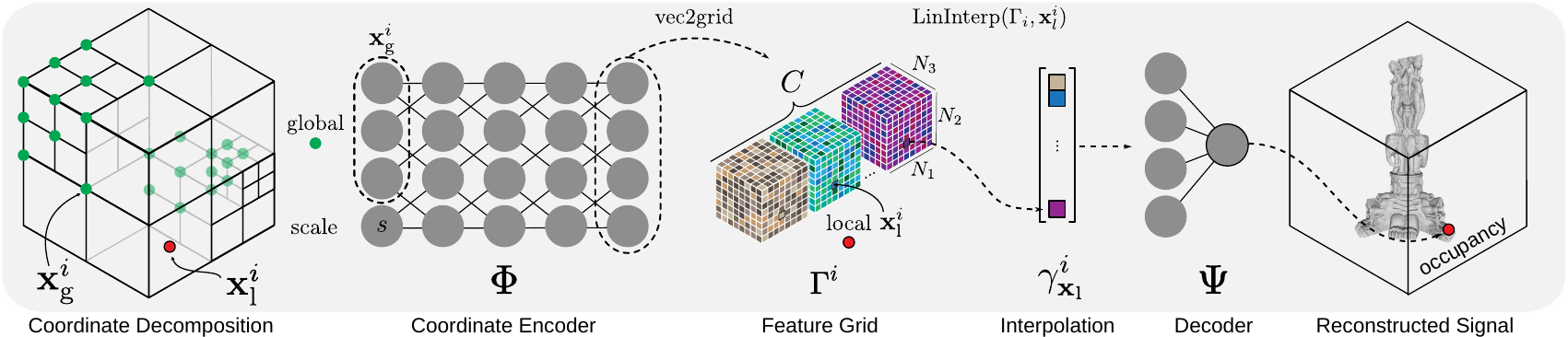}
\caption{\jm{Illustration of multiscale network architecture for 3D shape representation, including the adaptive coordinate decomposition, the larger coordinate encoder network, the discrete feature grid, feature interpolation, the smaller decoder network, and the reconstructed signal.}}
\label{fig:representation}
\end{figure*}
Our approach for multiscale representation networks relies on two main components: (1) a multiscale block parameterization that partitions the input space according to the local signal complexity, and (2) a network architecture consisting of a coordinate encoder and a feature decoder that efficiently map input spatial and scale coordinates to an output value.  




\subsection{Multiscale Block Parameterization}

At the core of our multiscale block parameterization is a tree-based partition of the input domain.
Specifically, we partition the domain with a quadtree in two dimensions or an \jm{octree} in three dimensions, and we fix the finest scale, or maximum depth, of the tree (see Figure~\ref{fig:representation}).
As opposed to traditional multiscale decompositions, where each value of the input domain is represented at multiple scales (e.g., each pixel in an image pyramid), our approach partitions space such that each input value is represented at a single scale.
That is, while a particular input domain value may fall within the bounds of potentially multiple quadrants or octants (hereafter \textit{blocks}) at different scales, we associate each value with one ``active'' block, $B_i$, at a single scale.
The input domain is thus partitioned into a set of active blocks, $\partition$, that together cover the domain. \jm{In summary, the ``active blocks'' are the currently selected blocks in the octree/quadtree that uniquely represent a region of the input space. The selection of active blocks is part of an optimization problem we describe in Section~\ref{subsec:online_decomp}.}

The blocks and input domain values within each block are addressed by a global block index and a continuous local coordinate.
The global block index, $\coarse^i\in\mathbb{R}^{\din+1}$, gives the global coordinate of $B_i$.
We define \jm{$\din$} as the dimensionality of the input coordinate domain and the last element of $\coarse^i$ encodes the normalized scale of the block corresponding to its discrete level in the quadtree or \jm{octree}.

We also introduce a continuous local coordinate $\fine^i$ to parameterize the input domain within $B_i$.
Here, $\fine^i$ is a $\din$-dimensional vector whose elements take on values between -1 and 1, or $\fine\in\mathbb{R}^{\din}_{[-1, 1]}=\{(x_1,\ldots,x_{\din})^T\ |-1 \leq x_i \leq 1 \}$.
Thus any continuous coordinate $\mathbf{x}$ in the input domain can be described by pairing the global block index, $\coarse^i$, of its containing block with the continuous local coordinate, $\fine^i$, which describes a relative position within the block.

\subsection{Neural Network Architecture}
The \jm{representation network} (see Figure~\ref{fig:representation}) contains a coordinate encoder 
\begin{equation}
 \encoder : \mathbb{R}^{\din + 1} \rightarrow \mathbb{R}^{C\times N_1\times\cdots\times N^{\din}}, \quad \coarse^i \mapsto \encoder(\coarse^i) = \Gamma^i, 
\end{equation}
which maps a global block index to a grid of features $\Gamma^i$ \jm{of dimension $N_1\times\cdots N_{\din}$ (i.e., it has size $N_i$ along the $i$-th dimension) with $C$ feature channels (i.e., each cell in $\Gamma^i$ is a vector of dimension $C$). Both the $N_i$'s and $C$ are hyperparameters.}
Given a continuous \jm{local coordinate $\fine$}, a feature vector is extracted as
\begin{equation}
    \jm{\interp\left(\Gamma^i, \fine^{i}\right) = \gamma^{i}_{\fine}\in\mathbb{R}^C,}
\end{equation}
where \jm{\interp\ is the multivariate linear interpolation (e.g., bilinear when $\din=2$ or trilinear when $\din=3$)} of the feature grid and is used to \jm{calculate} a feature $\gamma^{i}_{\fine}$ at \jm{any} relative coordinate $\fine^{i}$ within the grid.
A feature decoder $\decoder$ maps the feature vector to the output, $\out^{i}$, as 
\jm{
\begin{equation}
    \decoder :  \mathbb{R}^{C} \rightarrow \mathbb{R}, \quad \gamma_{\fine}^i \mapsto \decoder(\gamma_{\fine}^i) = \out^i \approx \out_{\gt}^i.
    \label{eq:decoder_output}
\end{equation}
}
Here, the output of the decoder approximates the signal of interest $\out_{\gt}$ at the global coordinate $\mathbf{x}$.

In practice, we implement both networks using multi-layer perceptrons (MLP) with ReLU nonlinearities.
We apply a positional encoding~\cite{mildenhall2020nerf} to the global block indices before \jm{input} to the coordinate encoder, and the feature decoder consists of a small MLP with a single hidden layer that is efficient to evaluate.

One key advantage of this two-stage architecture is that it greatly reduces the computational expense of evaluating many coordinates within the same block. 
While the coordinate encoder $\encoder$ contains the bulk of the network parameters and is the most expensive component to evaluate, only a single forward pass is required.
Once the feature grid is generated, all local coordinates within the block can be evaluated \jm{using a lightweight interpolation operation and a forward pass through} the small decoder network.
During training time, this also reduces memory requirements as we are only required to store intermediate activations of the larger coordinate encoder network (used for backpropagation) for a small number of block indices rather than every single input coordinate.
Another advantage is a potential improvement in network capacity since feature grids can be re-used for signals with repetitive structures across spatial locations and scales.

\subsection{Online Multiscale Decomposition}
\label{subsec:online_decomp}
One challenge in training neural representations efficiently is that most training samples should come from regions of the signal of interest that have the greatest complexity or are most challenging to learn. 
Likewise, at inference time, most of the computation should be devoted to accurately generating complex regions of the signal rather than simple regions.
For example, for an image, we expect that it is possible to learn and infer flat-textured regions more efficiently than highly textured regions while yet achieving the same image quality.
These observations are at odds with conventional neural representations which use an equal density of training samples for all regions of an image or 3D surface and require an equal number of forward passes through the network to render each region of the representation.

Here, we propose a new automatic decomposition method that allocates network resources adaptively to fit a signal of interest. 
Our technique is inspired by adaptive mesh refinement methods in simulation techniques and finite-element solvers which refine or coarsen meshes at optimization time to improve solution accuracy while minimizing computational overhead~\cite{berger1984adaptive}.

During training time, we fix the maximum number of active blocks in the decomposition, allocate $N_1\times\cdots\times N_{\din}$ training samples to each block, and track the average signal fitting loss within each block.
The maximum number of blocks is a hyperparameter that can be tuned according to the computational budget and network capacity.
At frequent intervals during training, we apply a non-differentiable update to the scale of each block by solving a resource allocation problem that seeks to minimize the fitting loss while maintaining fewer blocks than the prescribed maximum. 

Solving the resource allocation problem causes blocks to merge or combine, resulting in an update to the set of active blocks or global block indices used during training.
Since the number of training samples within each block is fixed, the procedure dynamically allocates more samples to harder-to-fit regions by placing finer blocks at the corresponding locations.
At inference time, coarse blocks are used to represent less detailed regions such that the signal can be generated with fewer forward passes through the more expensive coordinate encoder and more passes through the lightweight feature decoder.

\paragraph{Online optimization details.}

We formulate the resource allocation problem as an Integer Linear Problem (ILP) where each active block $B_i$ in the partition $\partition$ is considered in the context of its sibling blocks. 
In a quadtree, \jm{each node has four children}, thus each block, $B_i$ has a set of three siblings, $\mathcal{S}(i)$, and sibling groups are composed of $N_S=4$ blocks. 
Likewise, blocks in an \jm{octree} have seven siblings with $N_S=8$ blocks in a sibling group.
Thus, for any active block we can use three binary decision variables to encode whether it merges with its siblings to a coarser scale ($\imerge$), stays as is ($\inone$), or splits to create a new group of siblings (children) at a finer scale ($\isplit$):
\begin{equation}
 \bold{I_i}=\left(\imerge_i,\inone_i,\isplit_i\right)^\tp\in\{0,1\}^3.
\end{equation}

To merge a block, all siblings must be active (i.e., members of the current partition $\partition$) at the same scale and have $\imerge=1$.  We define a corresponding ``grouping'' indicator as 
\jm{
\begin{equation}
 \igrp_i \overset{\text{def.}}{=} \frac{1}{N_S}\left( \imerge_i + \sum_{j\in\mathcal{S}(i)} \imerge_j \right),
\end{equation}
}
which evaluates to $1$ when all siblings are set to merge. 

The decisions for a block to merge, split or stay at the same scale are mutually exclusive, which we encode in the constraint
\begin{equation}
 \igrp_i + \isplit_i + \inone_i = 1.
 \label{eq:cstr_choice}
\end{equation}

The partition obtained from the optimization must not contain more than $N_B$ blocks. If a block splits ($\isplit_i=1$) it contributes $N_S$ blocks to the new partition, or if it merges ($\imerge_i=1$) it contributes $\frac{1}{N_S}$ blocks. Thus, we can constrain the total number of blocks in the optimized partition as
\begin{equation}
 \left( \sum_i \frac{1}{N_S}\,\imerge_i + \inone_i + N_S\,\isplit_i \right) \leq N_B.
  \label{eq:cstr_knapsack}
\end{equation}

Ultimately we wish to optimize the block partition such that the new partitioning has a lower fitting error to the signal of interest after it is optimized. Thus, for each block, we require an estimate of the fitting error incurred from a decision to merge, stay the same, or split. We arrange these estimates into a vector of weights $\bold{w}_i=\left(w_i^{\uparrow},w_i^{=},w_i^{\downarrow}\right)^\tp$ and solve the following optimization problem to find the new partition.
\begin{align}	
 \text{minimize}\quad& \sum_i \bold{w}_i^\tp\,\bold{I}_i  \label{eq:knapsack} \\
 \text{subject to}\quad& 
 \begin{cases}
 \eqref{eq:cstr_choice}\: \forall B_i\in\mathcal{B}, \\
 \eqref{eq:cstr_knapsack}.
 \end{cases} \nonumber
\end{align}
To calculate the values of $\bold{w}_i$, we first compute the mean block error $\blockerr^i$ using an application dependent loss $\ell:\mathbb{R}^{\dout}\times\mathbb{R}^{\dout}\mapsto\mathbb{R}$ taken over the output of the feature decoder $\out^{i}$ for all the corresponding continuous local coordinates $\fine^{i}$.
\begin{equation}
    \blockerr^i = \mathbb{E}_{\fine}\left[ \ell\left(\out^{i},\out_{\gt}\left(\mathbf{x}\right)\right)\right].
\end{equation}
\jm{where $\mathbb{E}_{\fine}[\cdot]$ denotes the expectation (empirical mean) taken over the $N_l$ sampled fine coordinates $\fine$ (see Alg.~\ref{alg:training}).} Then, we calculate $w_i^{=}$ for each block with volume $\text{Vol}_i$ as
\begin{equation}
 w_i^{=} = \text{Vol}_i \cdot \blockerr^i,
 \label{eq:weight_stay}
\end{equation}
Here, weighing the mean block loss by the volume is important for a consistent error estimate across scales. 
The error for merging ($w_i^{\uparrow}$) is estimated as the error of the parent block, $\mathcal{P}(i)$, when an estimate of its error is available (that is, if the parent has been fitted in the past). Otherwise we approximate the error with $N_S$ times the current error. 
\begin{equation}
  w_i^{\uparrow} =
    \begin{cases}
      \frac{1}{N_S}\,w_{\mathcal{P}(i)}^=, & \text{if parent's error is available}, \\
      (N_S+\alpha)\, w_i^= & \text{otherwise}.
    \end{cases}
    \label{eq:weight_merge}
\end{equation}
Using $\alpha>0$ captures the assumption that the error when merging blocks is worse than $N_S$ times the current block error.

The error for splitting a block ($w_i^{\downarrow}$) is similarly estimated as the sum of the errors of the children blocks when those estimates are available. Otherwise, we use an approximate estimate of $\frac{1}{N_S}$ times the current error.
\begin{equation}
  w_i^{\downarrow} =
    \begin{cases}
      \sum_{j\in\mathcal{C}(i)} w_j^=, & \text{if children's errors available}, \\
      (\frac{1}{N_S}-\beta)\, w_i^= & \text{otherwise}.
    \end{cases}
    \label{eq:weight_split}
\end{equation}
with $\mathcal{C}(i)$ the children of $i$. Here, using $\beta>0$ assumes that the error when splitting a block is likely to be better than $\frac{1}{N_S}$ times the current error.
\begin{algorithm}[t!]
\caption{Adaptive Multiscale Network Training}
\label{alg:training}
\SetKwComment{Comment}{$\triangleright$\ }{}
\DontPrintSemicolon
\SetKwInOut{Parameters}{Parameters}
\Parameters{}
$T$: max number of iterations\;
$T_{\text{optim.}}$: number of iter. between each partition optimization\;
$T_{\text{pruning}}$: number of iter. between each partition pruning\;
\BlankLine
$t\gets 0 $ \;
\Repeat{$t\geq T$} {
	\ForEach{\textup{block $B^i\in \partition$ with global coord. $\coarse^i$}} {
		Compute the feature grid $\Gamma^i = \encoder(\coarse^i)$\;
		Stratified sample $N_l$ local coord. $\fine^{i,j}\sim[-1,1]^{\din},\forall j<N_l$\;
		\ForEach{local coord. $\fine^{i,j}$} {
			\jm{Calculate feature by interp. $\gamma^{i,j}_{\fine} = \interp\left(\Gamma_i,\fine^{i,j}\right)$}\;
			Decode the feature $\out^{i,j} = \decoder(\gamma^{i,j}_{\fine})$\;
		}
		
		Compute block error $\blockerr^i=\mathbb{E}_{\fine}[\ell(\out^{i,j},\out^{i,j}_{\gt})]$ \;
		Update ILP weights $\{w_i^{\uparrow},w_i^{=},w_i^{\downarrow}\}$ from $\blockerr^i$ \;
	}
	Compute loss $\mathcal{L}=\frac{1}{|\partition|}\sum_{B^i\in\partition} \blockerr^i$ \;
	Backpropagate $\mathcal{L}$ to update $\encoder$ and $\decoder$ \;
	\If{ $t \mod T_{\text{optim.}} = 0$} {
	Optimize the partition $\partition$ \;}
	\If{ $t \mod T_{\text{pruning}} = 0$} {
	Prune the partition $\partition$ \;
	}
	$t \gets t+1$
}
\end{algorithm}
\paragraph{Solving the ILP}
\jm{We solve the ILPs to optimality with a branch-and-bound algorithm using the solver implemented in the Gurobi Optimization Package \shortcite{gurobi}\footnote{free academic license}. The solver is fast: using a thousand active blocks takes less than $100$ ms which is negligible with respect to the training time of the networks since the optimization only occurs every few hundred training steps.
}

\paragraph{Interpretation}
\jm{The ILP selects blocks to split and merge based on the feedback of the error fitting signal. The linear constraint in Equation~\eqref{eq:knapsack} and the design of our weights Eq.~\eqref{eq:weight_stay},~\eqref{eq:weight_merge},~\eqref{eq:weight_split} act in a way that is intuitively similar to sorting the blocks based on the fitting error and splitting or merging blocks with the highest or lowest error. However, compared to such an approach, solving the ILP offers a few advantages. The main advantage of this formulation is that it allows us to only specify the maximum number of blocks in the partition, which we assumed correlates to the maximum capacity of our encoder. Besides, it avoids other heuristics such as specifying the number of blocks that can merge/split each iteration.}
\subsection{Pruning} 
As the partition gets optimized over the course of training, it happens that certain blocks represent regions that are perfectly uniform. Those regions can be described by a single value, for instance, the color of a constant area in an image, or zero for the empty space in an occupancy field. To avoid learning the same value for a whole region in the multiscale network, we prune the block from the partition. It cannot be further split and its value is set in a look-up table. Since the block is no longer active, it frees up space in the partition by loosening the constraint in Equation~\eqref{eq:knapsack}. Thus, the next time the ILP is solved, it can further split active blocks. This will result in a finer partition. To decide whether to prune a block $B_i$ we found that meeting both the following criteria worked very well in practice:
\begin{description}
	\item[Low error] we check that the fitting error $\blockerr_i$ of the block $B_i$ is less than an application specific threshold (depending on the choice of $\ell(\cdot,\cdot)$).
	\item[Low variance in the block] we make sure that the values output by the network $\{\decoder(\fine^{i,j})\}_j$ in the block are not spread. 
\end{description}
\highlight{We show results of pruning in a following section on representing 3D scenes.} 

\section{Representing Gigapixel Images}
\label{sec:images}
\begin{figure*}
    \includegraphics[trim=0cm 8cm 0cm 0cm,clip]{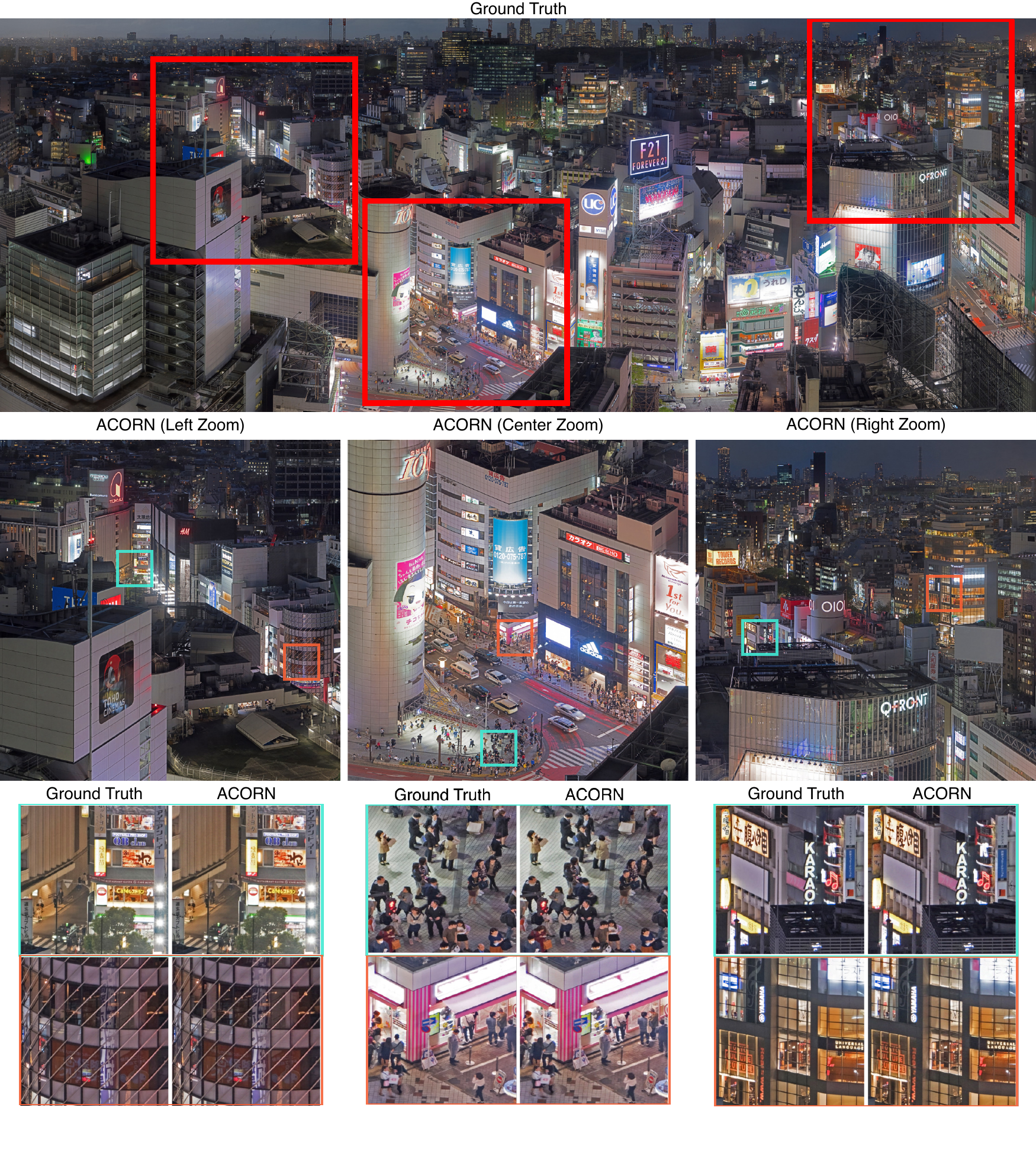}
    \caption{Gigapixel-scale image fitting. We show the results of fitting a Gigapixel panorama photo of Tokyo using the proposed multiscale architecture, achieving a PSNR of 38.59. Top row; ground truth input. Middle row; 10K $\times$ 10K zoom-ins of multiscale result. Bottom row; 1K $\times$ 1K zoom-ins of ground truth and multiscale result. (Photo: Trevor Dobson, \url{https://creativecommons.org/licenses/by-nc-nd/2.0/)}}
		\label{fig:tokyo}
\end{figure*}
In this section, we discuss the large-scale image representation experiments we show in Figures~\ref{fig:pluto} and~\ref{fig:tokyo}. Note that previous neural image representations (e.g.,~\cite{sitzmann2020siren,chen2020learning}) have been limited to images with resolutions of up to a megapixel (MP). With resolutions of 64~MP and $\sim$1~GP, respectively, our experiments far exceed these limits. 
\begin{table}[t]
\centering
\caption{Quantitative image fitting results.}
\label{tbl:image_quant}
\resizebox{\columnwidth}{!}{
\begin{tabular}{lcccccccc}
\toprule
& Train Time & GPU Mem. & Model Param. & PSNR (dB) $\uparrow$  & SSIM $\uparrow$ \\
\midrule
\multicolumn{6}{c}{Pluto Image (64 MP)} \\
\midrule
    ReLU P.E. & 33.9 h & 13.3 GB & 9.5 M & 35.99 &  0.93 \\
    SIREN  & 37.0 h & 20.9 GB & 9.5 M & 35.08 & 0.92\\
    ACORN (fixed) & 2.6 h & 2.4 GB & 9.5 M & 38.08 & 0.95 \\
    ACORN & \textbf{1.8 h} & \textbf{2.4 GB} & 9.5 M & \textbf{41.90} & \textbf{0.97}\\
\midrule
\multicolumn{6}{c}{Tokyo Image (996 MP)} \\
\midrule
    ACORN & 36.9 h & 33.0 GB & 168.0 M & 38.59 & 0.94 \\
\bottomrule
\end{tabular}}
\vspace{-1em}
\end{table}
\subsection{Image Fitting Task}

To fit an image with ACORN, we train in an online fashion where we generate $N_l$ random stratified samples within active block of the multiscale decomposition.
Then, the values of $\out_{\gt}$ are determined by bilinear interpolation of the corresponding global coordinates on the target image. 
The training proceeds according to Alg.~\ref{alg:training}, though we disable the pruning step since we assume all blocks contain some non-zero texture.
We use a mean-squared error loss between the network output and the target image for each block as  
\begin{equation}
    \ell(\out^i, \out^i_{\gt}) = \lVert \out^i - \out^i_{\gt} \rVert_2^2.
\end{equation}
During training, we optimize the block partition every 500 iterations and fix the maximum number of blocks to $N_B=1024$. 

\paragraph{Baselines.}
We compare our method to state-of-the-art coordinate representation networks designed for fitting images: sinusoidal representation networks (SIREN), and an MLP using ReLU non-linearities and a frequency-based positional encoding strategy~\cite{mildenhall2020nerf,tancik2020fourier}.
We also add an additional ablation study that disables the multiscale block optimization and instead uses a fixed decomposition by evenly distributing the $N_B$ blocks at a single scale across the domain.
To compare the methods we set the parameter counts of the corresponding networks to be approximately equal. 
More details about the optimization and network architectures are provided in Appendix~\ref{sec:implementation}. 

\subsection{Performance}

We demonstrate the performance of ACORN by fitting two large-scale images.  
The first is an image of Pluto (shown in Figure~\ref{fig:pluto}), captured by the New Horizons space probe.
The image has a resolution of 8,192 $\times$ 8,192 pixel (64 MP) and has features at varying scales, making it ideal for a multiscale representation. 
The adaptive grid optimized by our resource allocation strategy clearly shows that the representation uses significantly smaller blocks to represent fine details, such as the craters, whereas the empty and large uniform areas are represented at much coarser scales.
As seen in the convergence plots, our representation converges about two orders of magnitude faster compared to state of the art coordinate networks and it fits this challenging image at a significantly higher quality well above 40~dB PSNR (see Table~\ref{tbl:image_quant} and color-coded error maps in Figure~\ref{fig:pluto}). 
All baselines are evaluated on the same an NVIDIA RTX 6000 GPU, and metrics are reported after 100,000 iterations of training.

In addition to fitting this image faster and better than alternative network architectures, our model only requires 2.4~GB of memory whereas ReLU with positional encoding (P.E.)~\cite{mildenhall2020nerf} and SIREN~\cite{sitzmann2020siren} require 13.3~GB and 20.9~GB, respectively. 
Note that while the parameter counts of the methods are similar, our approach uses far less memory because of the two-stage inference technique.
For backpropagation, intermediate activations through the coordinate encoder are stored only once per block while intermediate activations for the feature decoder (with far fewer parameters) are stored for each local coordinate.
\begin{table}[t!]
\caption{Comparison of ACORN models fitted on the Pluto image ($1024\times 1024$) varying the number of blocks used in the quadtree decomposition (training time reported using a NVIDIA Titan V).}
\label{tbl:varying_num_blocks}
\jm{
\begin{tabular}{lcccc}
\toprule
Number of blocks & 64 & 256 & 1024 & 4096 \\
\midrule
PSNR at convergence (dB) & 31.4 & 37.3 & 40.1 & 37.7 \\
Training time to 50 K it. (min) & 45 & 63 & 140 & 388 \\
Training time to 30 dB PSNR (min) & 7 & 4 & 10 & 48 \\
\bottomrule
\end{tabular}
}
\vspace{-1em}
\end{table}

We also note the improved performance of our technique relative to using the same network architecture, but with a fixed block decomposition.
While fixing the block decomposition yet results in rapid convergence and a network that outperforms other methods (see Figure~\ref{fig:pluto} and Table~\ref{tbl:image_quant}), we find that including the block decomposition significantly improves image quality. 
We also notice slightly faster training times using the adaptive block decomposition because we start the training with only 64 blocks at a coarse scale, leading to faster iterations at the beginning of training.

The gigapixel image of Tokyo shown in Figure~\ref{fig:tokyo} has a resolution of 19,456 $\times$ 51,200 pixels, which is about three orders of magnitude higher than what recent neural image representations have shown. 
Again, this image captures details over the entire image at different scales. The multiscale architecture is the same as the one used to fit the Pluto result in Figure~\ref{fig:pluto}, though we increase the parameter count as described in Table~\ref{tbl:image_quant} and Appendix~\ref{sec:implementation}. A visualization of the optimized grid showing the block decomposition and additional training details are included in the supplement. 

As we have shown, ACORNs can be scaled to flexibly represent large-scale 2D images, offering significant improvements to training speed, performance, and the overall practicality of this task.

\subsection{Interplay between the quadtree decomposition, time and fitting accuracy}
\jm{We perform an experiment that varies the number of blocks used in the fitting of the Pluto image (Figure~\ref{fig:pluto}) downsampled at a resolution of $1024\times 1024$. Results are summarized in Table~\ref{tbl:varying_num_blocks}. With fewer blocks, the image is represented at a coarser level of the octree: training is faster since there are fewer blocks to fit but the overall model accuracy is lower. Increasing the number of blocks improves accuracy until the coordinate encoder saturates and fails to generate accurate features across the entire image at the finest level of the octree.}

\section{Representing Complex 3D Scenes}
\label{sec:threed}
%
\begin{table}[t!]
\centering
\caption{Quantitative 3D fitting results (Thai Statue).}
\label{tbl:thai_quant}
\resizebox{\columnwidth}{!}{
\begin{tabular}{lcccccccc}
\toprule
& Train Time & \multicolumn{2}{c}{GPU Mem.} & \multicolumn{2}{c}{Model Param.} \\
\midrule
SPSR 		& 0.128 h & \multicolumn{2}{c}{N/A}    & \multicolumn{2}{c}{N/A} \\
Conv. Occ.  & 100 h   & \multicolumn{2}{c}{9.1 GB} & \multicolumn{2}{c}{4.167 M}\\
SIREN 		& 29.28 h & \multicolumn{2}{c}{14.3 GB}& \multicolumn{2}{c}{16.8 M} \\
ACORN  		& 4.25 h  & \multicolumn{2}{c}{1.7 GB} & \multicolumn{2}{c}{17.0 M} \\
\midrule
& IoU ($\uparrow$) & Chamfer-$L_1$ ($\downarrow$) & Precision ($\uparrow$) & Recall ($\uparrow$) & F-Score ($\uparrow$) \\
\midrule
SPSR 		&  0.809 & 4.76e-04 & 0.891 & 0.897 & 0.894 \\
Conv. Occ.  &  0.965 & 2.16e-05 & 0.972 & 0.992 & 0.982  \\
SIREN 		& 0.992  & 2.06e-05 & 0.996 & 0.996 & 0.996 \\
ACORN  		& {\bfseries 0.999} & {\bfseries 1.55e-05} & {\bfseries 0.999} & {\bfseries 0.999} & {\bfseries 0.999} \\
\bottomrule
\end{tabular}}
\vspace{-1em}
\end{table}
%

%
\begin{table}[t!]
\centering
\caption{Quantitative 3D fitting results (Engine).}
\label{tbl:engine_quant}
\resizebox{\columnwidth}{!}{
\begin{tabular}{lcccccccc}
\toprule
& Train Time & \multicolumn{2}{c}{GPU Mem.} & \multicolumn{2}{c}{Model Param.} \\
\midrule
SPSR 		& 0.114 h &  \multicolumn{2}{c}{N/A} 	& \multicolumn{2}{c}{N/A} \\
Conv. Occ.  & 108 h   &  \multicolumn{2}{c}{9.1 GB} & \multicolumn{2}{c}{4.167 M} \\
SIREN 		& 29.21 h &  \multicolumn{2}{c}{14.3 GB}& \multicolumn{2}{c}{16.8 M} \\
ACORN		& 7.77 h  &  \multicolumn{2}{c}{1.7 GB} & \multicolumn{2}{c}{17.0 M} \\
\midrule
& IoU ($\uparrow$) & Chamfer-$L_1$ ($\downarrow$) & Precision ($\uparrow$) & Recall ($\uparrow$) & F-Score ($\uparrow$) \\
\midrule
SPSR 		& 0.843 & 1.19e-04 & 0.906 & 0.924 & 0.951 \\
Conv. Occ.  & 0.840 & 7.93e-04 & 0.877 & 0.952 & 0.913 \\
SIREN 		& 0.961 & 1.09e-04 & 0.980 & 0.980 & 0.980 \\
ACORN	    & {\bfseries 0.987} & {\bfseries 8.46e-05} & {\bfseries 0.992} & {\bfseries 0.994} & {\bfseries 0.993} \\
\bottomrule
\end{tabular}}
\vspace{-1.6em}
\end{table}
\begin{figure*}
    \includegraphics[scale=0.97,trim=0cm 0cm 0cm 0cm, clip]{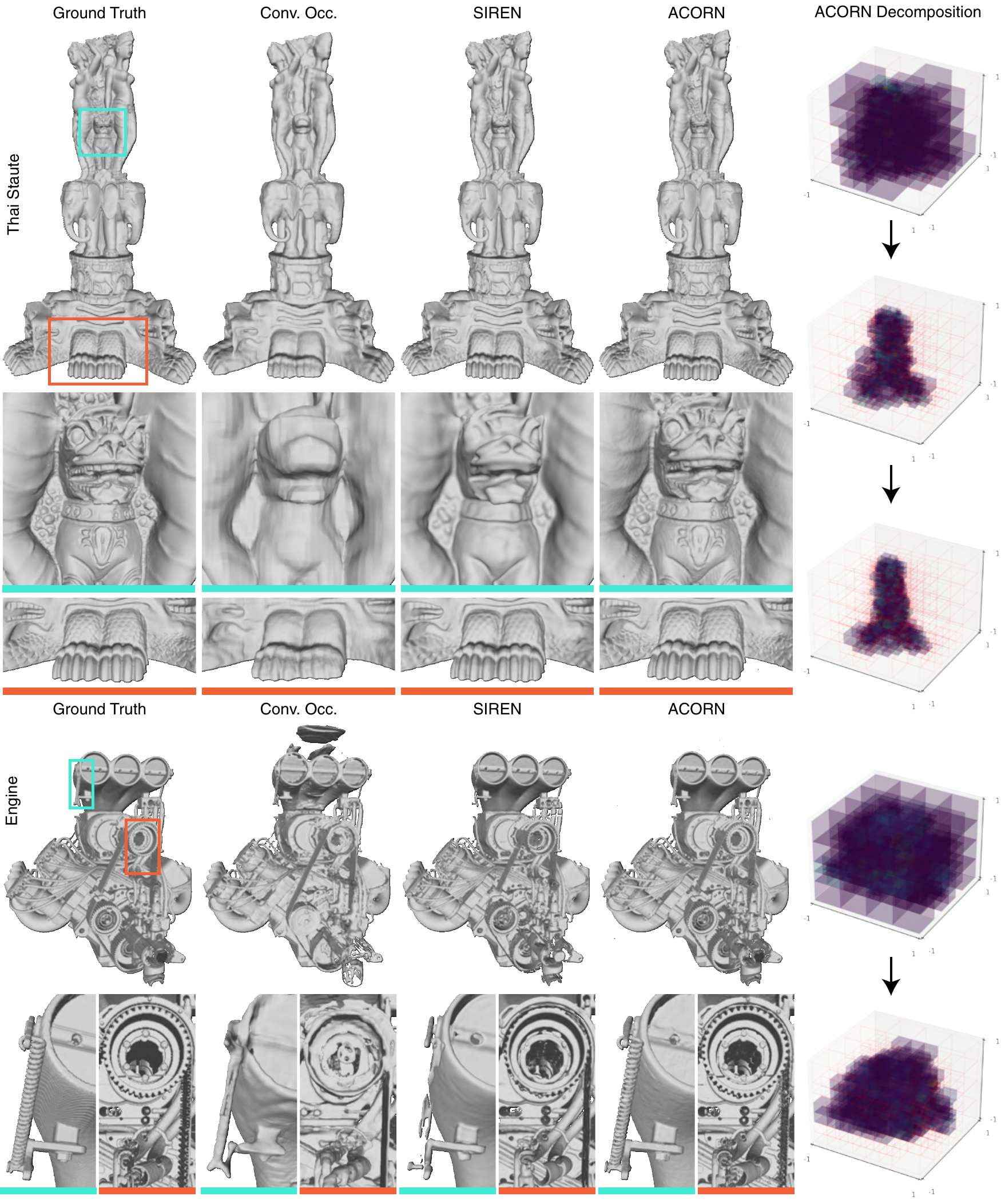}
    \vspace{-0.25cm}
    \caption{Complex 3D scene fitting. We compare the results of fitting highly detailed meshes of a Thai statue and an engine using the baseline methods and our proposed multiscale architecture. ACORN is able to capture both global and local geometric features better than all other baselines. \highlight{The evolution of the learned block decompositions during training are shown on the right.} (Engine Model: ChrisKuhn, \url{https://creativecommons.org/licenses/by/2.0/})} 
	\label{fig:thai_engine}
\end{figure*}

The benefits of a multiscale representation follow when moving from image representation to 3D scene representation. We benchmark our method against Screened Poisson Surface Reconstruction (SPSR)~\cite{kazhdan2013screened}, a classical method for obtaining watertight surfaces from point clouds, Convolutional Occupancy Networks~\cite{peng2020convolutional}, a recent example of local shape representations, and SIREN~\cite{sitzmann2020siren}, a state-of-the-art coordinate representation.

We compare representations on the task of representing the occupancy field of detailed 3D models. The Thai Statue, obtained from the Stanford 3D Scanning Repository \shortcite{stanford}
, consists of 10 million polygons and depicts an intricate resin statue. While the general shape is simple, fine details of the reliefs are challenging to capture with existing methods. The Dragster Engine, obtained free from BlendSwap, consists of 512 thousand polygons and is a recreation of an engine used in Top Fuel drag-racing. The thin features and hard edges that make up the engine are difficult to faithfully recreate.

\subsection{Occupancy Prediction Task}

We benchmark representations on the task of fitting the occupancy fields~\cite{mescheder2019occupancy} of 3D models. At training time, we sample 3D points $\bold{x} \in \mathbb{R}^3$ with a 50/50 split -- half of the points are sampled near the mesh surface, half of the points are sampled uniformly within the volume. Given the input points, we predict occupancy values. We apply binary cross-entropy loss between the occupancy prediction of our network, $\out\in\mathbb{R}$ and the true occupancy, $\out_{\gt}$:
\begin{equation}
	\mathcal \ell(\out, \out_{\gt}) = -[\out_{\gt} \cdot \text{log}(\out) + (1 - \out_{\gt}) \cdot \text{log}(1 - \out)].
\end{equation}

As in \cite{mescheder2019occupancy}, we compute the ground truth occupancy of arbitrary points using the even--odd rule. In contrast to the deep-learning approaches, SPSR instead takes a point cloud with oriented normals as input.

\subsection{Applying Multiscale Model to Occupancy Prediction}

In moving from representing images in two dimensions to representing shapes in three dimensions, we replace the quadtree with an \jm{octree} and predict a dense feature cube rather than a feature grid. As described above, our online multiscale decomposition approach samples coordinates per octant, which dynamically allocates samples towards hard-to-fit regions and improves the sample efficiency of training. Details about the network architecture can be found in \highlight{Appendix~\ref{sec:implementation}}.

\paragraph{Empty Octant Pruning}

The nature of our multiscale representation, which represents the scene at different scales, enables an additional optimization to further improve training time and performance. Periodically during training, we identify octants that are likely to be empty and ``prune'' them from the tree. In doing so, we eliminate sampling from regions of empty space while also freeing up octants and model capacity for more difficult and complex regions \highlight{(see Figure~\ref{fig:thai_engine})}.

\subsection{Metrics}
We evaluate the performance of all models by predicting a $1024^3$ cube of occupancy predictions. We compare against the ground-truth occupancy values with Intersection over Union (IoU), Precision, Recall, and F-Score. We additionally use marching cubes \cite{lorensen1987marching} to extract surfaces and evaluate mesh similarity with Chamfer distance. Quantitative results for these metrics, along with the model size, GPU memory required at training, and training time required to reach the reported results, are reported in Tables~\ref{tbl:thai_quant} and~\ref{tbl:engine_quant}.
All experiments are run on an RTX 6000 GPU.

\subsection{Evaluating Performance}

\highlight{As shown in Figure~\ref{fig:thai_engine}, ACORN} represents complex shapes with higher accuracy than previous methods. Qualitatively, it is more capable than previous implicit methods (Conv. Occ., SIREN) at representing fine details, such as intricate reliefs and tightly-coiled springs. Quantitatively, our approach outperforms all baselines on volume and mesh accuracy metrics.

When compared to previous neural representations, the improvements in computational efficiency become apparent. By sharing the computation among sampled points, our approach significantly reduces the memory and time costs associated with training and querying a model. When trained with the same number of parameters and at the same batch size, our multiscale representation requires nearly an order of magnitude less memory than a comparable MLP-based implicit representation, and yet outperforms it on representation accuracy metrics.

\section{Compression}
\label{sec:compression}
\jm{
Memory usage presented in Tables~\ref{tbl:image_quant},~\ref{tbl:thai_quant}~and~\ref{tbl:engine_quant} corresponds to GPU memory needed to store and train ACORN. The size of the representation itself is much smaller: it is equal to the number of parameters multiplied by their encoding size (32-bits floats in all our experiments). 
In Table~\ref{tbl:compression_quant} we provide the size of the uncompressed ACORN models (i.e., no weight quantization nor pruning). For context we also provide the uncompressed sizes of the images and 3D models. The network representation wins by a large margin against the uncompressed models in most cases. Compression could be applied to the ACORN representions in different ways to gain further savings (e.g., using pruning and quantization~\cite{han2015deep}).
}
\begin{table}
\caption{Size of ACORN models and of the uncompressed and compressed data used for fitting.}
\label{tbl:compression_quant}
\resizebox{\columnwidth}{!}{
\jm{
	\begin{tabular}{clccc}
	\toprule
	&							& Uncompressed & Compressed & ACORN \\
	\midrule
	\parbox[t]{2mm}{\multirow{2}{*}{\rotatebox[origin=c]{90}{img.}}}					
	&Pluto (8192$\times$8192)		& n/a & 192~MB & 38~MB \\
	&Tokyo (19456$\times$51200)		& 2.8~GB & 169~MB & 670~MB \\ \midrule
	\parbox[t]{2mm}{\multirow{4}{*}{\rotatebox[origin=c]{90}{shapes}}}	
	&Lucy (14~M vert.)		& 1.2~GB & 380~MB & 68~MB\\
	&Dragon (930~K vert.)	& 66~MB & 27~MB & 68~MB\\
	&Thai Statue (5~M vert.)	& 424~MB & 126~MB & 68~MB \\
	&Engine (308~K vert.)		& 16~MB 	& 4.4~MB & 68~MB \\
	\bottomrule
	\end{tabular}
}
}
\vspace{-1em}
\end{table}

%
%
%
%
%
%

\section{Discussion}
\label{sec:discussion}
In summary, we propose an adaptive multiscale neural scene representation that fits large-scale 2D and complex 3D scenes significantly faster and better than existing neural scene representations. 
These capabilities are enabled by a new hybrid implicit--explicit coordinate network architecture and a custom training routine that adapts the capacity of the network to the target scene in an optimized multiscale fashion.

\paragraph{Limitations}

\begin{figure}[t]
\includegraphics[]{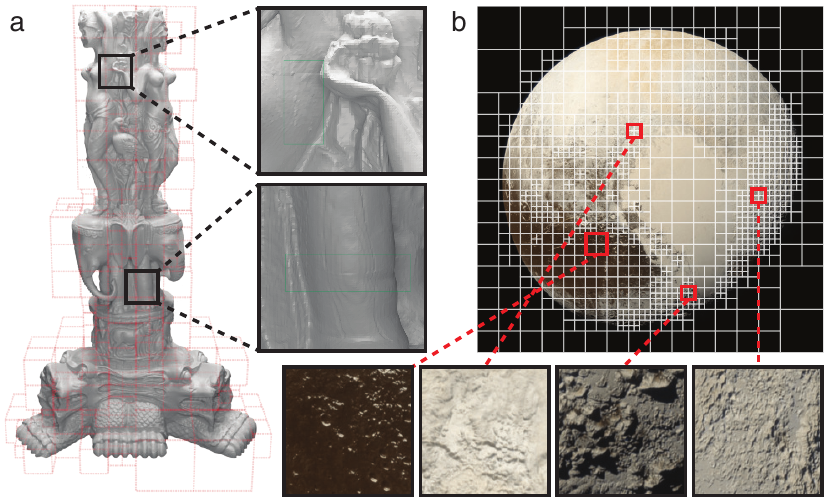}
	\caption{Demonstration of fitting at block borders. Though we do not explicitly enforce continuity across blocks, the borders between blocks yet retain high fitting quality. However, it is possible to observe small artifacts at the borders in the shape fitting (a) and image fitting (b) examples.}
\label{fig:seams}
\end{figure}

Currently, our method is limited in several ways.
First, each block of our decomposition represents the scene at a single scale rather than multiple scales to preserve network capacity.
The selected decomposition typically represents the coarsest scale at which a specific scene part can be represented adequately.
Yet, it might be helpful to represent scenes at multiple scales simultaneously.
For example, this would provide an option to evaluate signals faster at coarser resolutions, \highlight{similar to recent work~\cite{takikawa2021neural}}.
Second, although the integer linear program we use to update the block partitioning integrates well with our training routine, it is non-differentiable.
Exploring differentiable alternatives may further accelerate convergence and efficiency of our method.
Third, updating the partitioning requires retraining on new blocks, so we see temporary increases in the fitting loss periodically during training.
Again, a differentiable form of the resource allocation could alleviate these shortcomings.
Fourth, the maximum number of blocks is fixed a priori rather than globally optimized, due to the formulation of the integer linear program. 
\highlight{Finally, we note that the training method does not explicitly enforce continuity of the signal across blocks, though we train the representation on coordinates within the entirety of each block (including the boundary). 
In practice, we observe high-quality fitting near the boundaries, though some minor artifacts can occasionally be observed (see Fig.~\ref{fig:seams}).}
%

\paragraph{Future Work}

In future work, we would like to address the aforementioned limitations.
Moreover, we would like to explore additional applications in graphics and interactive techniques, such as neural rendering or fluid simulation.
\jm{Source code is publicly available to foster application-domain-specific adjustments of our representation for any of these applications.}
Neural scene representations in general offer new ways to compress scenes, for example by quantizing or pruning the network after training, which would be interesting to explore in the future.
We would also like to explore level-of-detail queries, e.g., for faster rendering at lower resolutions and explore representing higher-dimensional scenes, such as spatio-temporal data or graph based models.


\section{Conclusion}

Rapidly emerging techniques at the intersection of neural computing and computer graphics are a source of new capabilities in applications related to rendering, simulation, imaging, and geometric modeling. 
For these techniques, the underlying signal representations are crucial, but have been limited in many respects compared to traditional graphics representations. With our work, we take steps towards closing this gap by enabling large-scale images and complex shapes to be efficiently represented by neural networks.

\section*{Acknowledgments}
J.N.P. Martel was supported by a  Swiss  National  Foundation  (SNF) Fellowship (P2EZP2 181817).
C.Z. Lin was supported by a David Cheriton Stanford Graduate Fellowship.
G.W. was supported by an Okawa Research Grant, a Sloan Fellowship, and a PECASE by the ARO. Other funding for the project was provided by NSF (award numbers 1553333 and 1839974).

%
%
%
%

\bibliographystyle{ACM-Reference-Format}
\bibliography{references}

\appendix
\section{Implementation Details}
\label{sec:implementation}
\subsection{Image Fitting}
To fit the Pluto image we train our ACORN using a ReLU multi-layer perception (MLP) for the coordinate encoder and the feature decoder. The coordinate encoder is composed of 4 hidden layers with 512 hidden units each, and the final hidden layer outputs a feature grid with $C=16$ channels and width and height dimensions of $N_1=N_2=32$. We normalize the global block indices to fall within -1 and 1 and pass them through the positional encoding layer described by Mildenhall et al.~\shortcite{mildenhall2020nerf} with 6 frequencies. The feature decoder uses a single hidden layer with 64 hidden units. We set the maximum number of blocks to $N_B=1024$ and optimize for 100,000 iterations using the Adam optimizer~\cite{kingma2014} and a learning rate of $1\times 10^{-3}$. The block partition is optimized every 500 iterations \highlight{with $\alpha=0.2$ and $\beta=0.02$}. For the baseline experiments with SIREN~\cite{sitzmann2020siren} and ReLU P.E.~\cite{mildenhall2020nerf}, we use MLPs with 4 hidden layers and 1536 hidden units to match the parameter count of the ACORN. 

For the gigapixel image of Tokyo, we use the same architecture, but adjust the number of hidden units and layers. Here, we use $C=64$, $N_1=19$, and $N_2=50$. The coordinate encoder uses 10 positional encoding frequencies with 8 hidden layers, each with 2048 hidden units. Here, the block sizes are set to preserve the aspect ratio of the $19456\times 51200$ resolution image. The feature decoder is kept the same as previously described and the maximum number of blocks is $N_B=16384$. We optimize the image for 57,000 iterations, at which point the loss appeared to be converged. 

\subsection{3D Occupancy Prediction}
ACORN for 3D representations uses ReLU MLPs for the coordinate encoder and the feature decoder. The coordinate encoder is composed of 4 hidden layers with 512 hidden units each, and the final hidden layer outputs a feature cube of $C=18$ channels and width/height/depth dimensions of $N_1=N_2=N_3=12$. We normalize the global block indices to fall within -1 and 1 and pass them through the positional encoding layer described by Mildenhall et al.~\shortcite{mildenhall2020nerf} with 6 frequencies. The feature decoder uses a single hidden layer with 64 hidden units. We set the maximum number of octants to $N_B=1024$. We optimize over 48,000 iterations for the Thai Statue and 84,000 iterations for the Engine, using the Adam optimizer~\cite{kingma2014} and a learning rate of $1\times 10^{-3}$. We sample 64 points per octant per iteration, for a total batch size of roughly 66k points per batch. Pruning is performed every 4000 iterations, and the block partition is optimized every 1000 iterations \highlight{with $\alpha=0.2$ and $\beta=0.02$}, unless pruning is performed. \highlight{Octants are pruned if the average sample error is less than $1 \times 10^{-3}$ and if the maximum predicted occupancy value is less than $5\times 10^{-5}$.} The total trainable parameter count for ACORN was roughly 17 million parameters.

\section{Additional Quantitative Results}
\label{sec:addl_results}
\subsection{Teaser figure: ``Lucy" and ``Dragon"}
\jm{We present the evaluation for the Lucy and Dragon three-dimensional models presented in Figure~\ref{fig:teaser} reconstructed with ACORN and with a SIREN baseline in Table~\ref{tbl:teaser_quant}.}
\begin{table}[hb!]
\caption{Quantitative evaluation of ACORN vs. the SIREN baseline on the shapes fitted in the teaser.}
\label{tbl:teaser_quant}
\jm{
	\resizebox{\columnwidth}{!}{\begin{tabular}{llccccc}
	\toprule
		& & IoU ($\uparrow$) & Chamfer-$L_1$ ($\downarrow$) & Precision ($\uparrow$) & Recall ($\uparrow$) & F1 ($\uparrow$) \\
	\midrule
	\parbox[t]{2mm}{\multirow{3}{*}{\rotatebox[origin=c]{90}{\small Dragon}}}		
		&Conv. Occ. & 0.961 & 3.98e-05 & 0.933 & 0.961 & 0.947 \\
		&SIREN & 0.980 &  1.18e-05 & 0.990 & 0.990 & 0.990 \\
		&ACORN & 0.994 &  1.04e-05 & 0.997 & 0.996 & 0.997 \\
	&&&&& \\
	\parbox[t]{2mm}{\multirow{3}{*}{\rotatebox[origin=c]{90}{\small Lucy}}} 	
	&Conv. Occ. & 0.961 & 2.27e-05 & 0.981 & 0.980 & 0.980 \\
		&SIREN & 0.993 & 1.06e-05 & 0.996 & 0.997 & 0.997 \\
		&ACORN & 0.997 & 1.17e-05 & 0.999 & 0.998 & 0.999 \\
	\bottomrule
	\end{tabular}}
}
\end{table}

\end{document}


\title{{\textsc acorn}: Adaptive Coordinate Networks for Neural Scene Representation\\Supplemental Information}

\author{Julien N. P. Martel*, David B. Lindell*, Connor Z. Lin, Eric R. Chan, Marco Monteiro \lowercase{and} Gordon Wetzstein}
\affiliation{%
  \institution{Stanford University}
  \city{Stanford}
  \state{CA}
  \country{USA}
}
\email{jnmartel@stanford.edu}
\email{lindell@stanford.edu}
\thanks{*~J.~N.~P. Martel and D.~B. Lindell equally contributed.}

\renewcommand\shortauthors{Martel, J. N. P. and Lindell, D. B. et al.}
%
\newcommand{\jm}[1]{#1}
\newcommand{\highlight}[1]{#1}
\newcommand{\gw}[1]{#1}
\newcommand{\ec}[1]{#1}

\newcommand{\interp}{\textsc{LinInterp}}
\newcommand{\coarse}{\mathbf{x}_\text{g}}
\newcommand{\fine}{\mathbf{x}_\text{l}}
\newcommand{\encoder}{\Phi}
\newcommand{\decoder}{\Psi}
\newcommand{\out}{\bold{y}}
\newcommand{\gt}{\text{GT}}
\newcommand{\din}{d_{\text{in}}}
\newcommand{\dout}{d_{\text{out}}}
\newcommand{\blockerr}{E}

\newcommand{\imerge}{I^{\uparrow}}
\newcommand{\isplit}{I^{\downarrow}}
\newcommand{\inone}{I^{=}}
\newcommand{\igrp}{I^{g}}
\newcommand{\tp}{\intercal}
\newcommand{\partition}{\mathcal{B}}

\maketitle
\thispagestyle{empty}

\pagestyle{plain}
\begin{figure*}[t!]
    \includegraphics[width=6.5in]{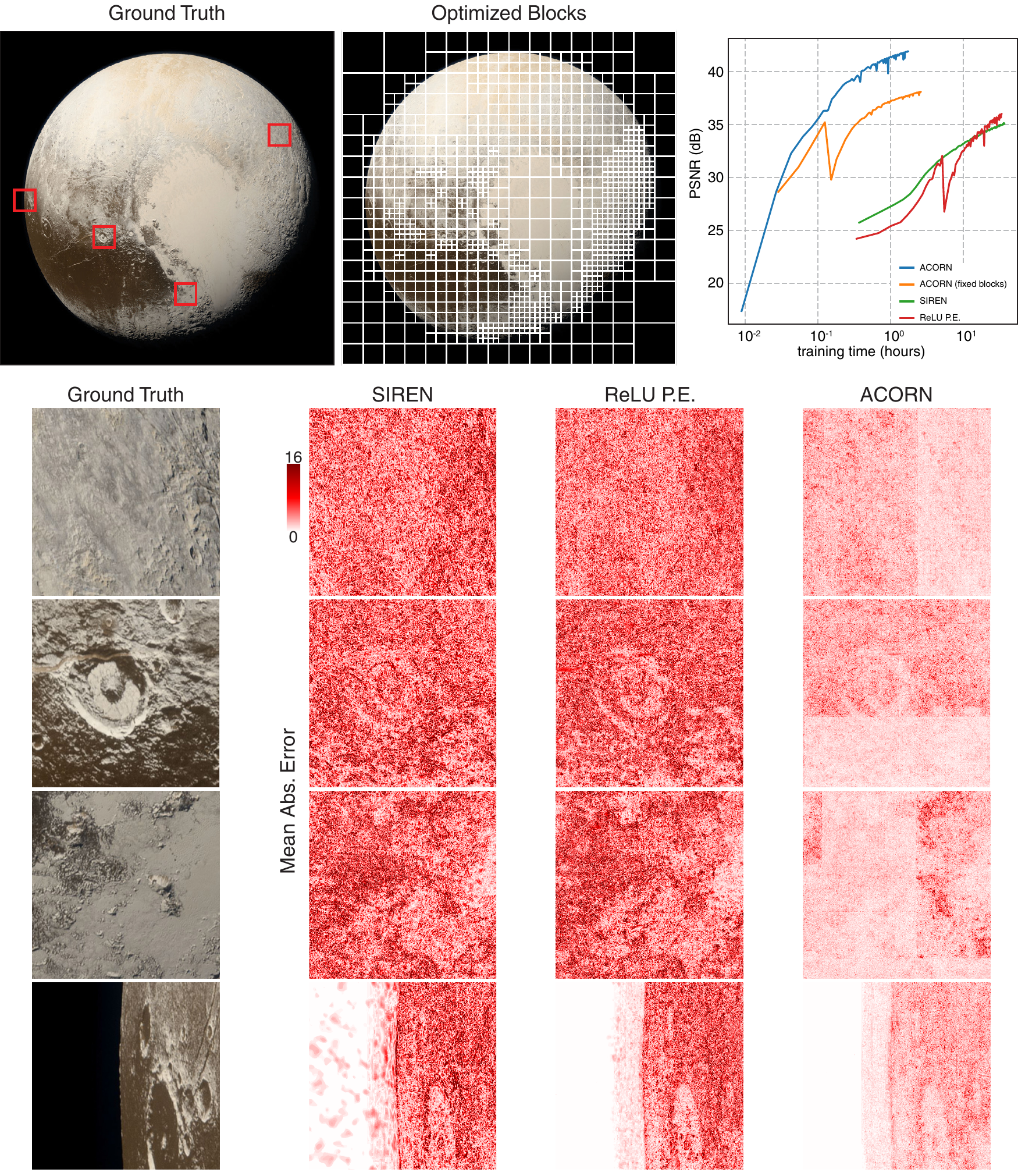}
    \caption{Large-scale image fitting. We fit networks to a 64 MP image of the dwarf planet Pluto, captured by the New Horizons space probe. The optimized block decomposition for our method is shown along with error maps for an inset of the image for our method, as well as a ReLU P.E. network~\cite{mildenhall2020nerf} and a SIREN~\cite{sitzmann2020siren}. All networks use roughly 9.5 M parameters. A plot of image PSNR vs. training time for all networks (including ours with a fixed, uniform block decomposition) is shown from 1,000 to 100,000 iterations. Our multiscale representation network converges to a PSNR of >40~dB in about one hour, whereas other representations struggle to adequately fit this image even more than one day.}
		\label{fig:supp_pluto}
\end{figure*}
\begin{figure*}
    \includegraphics[]{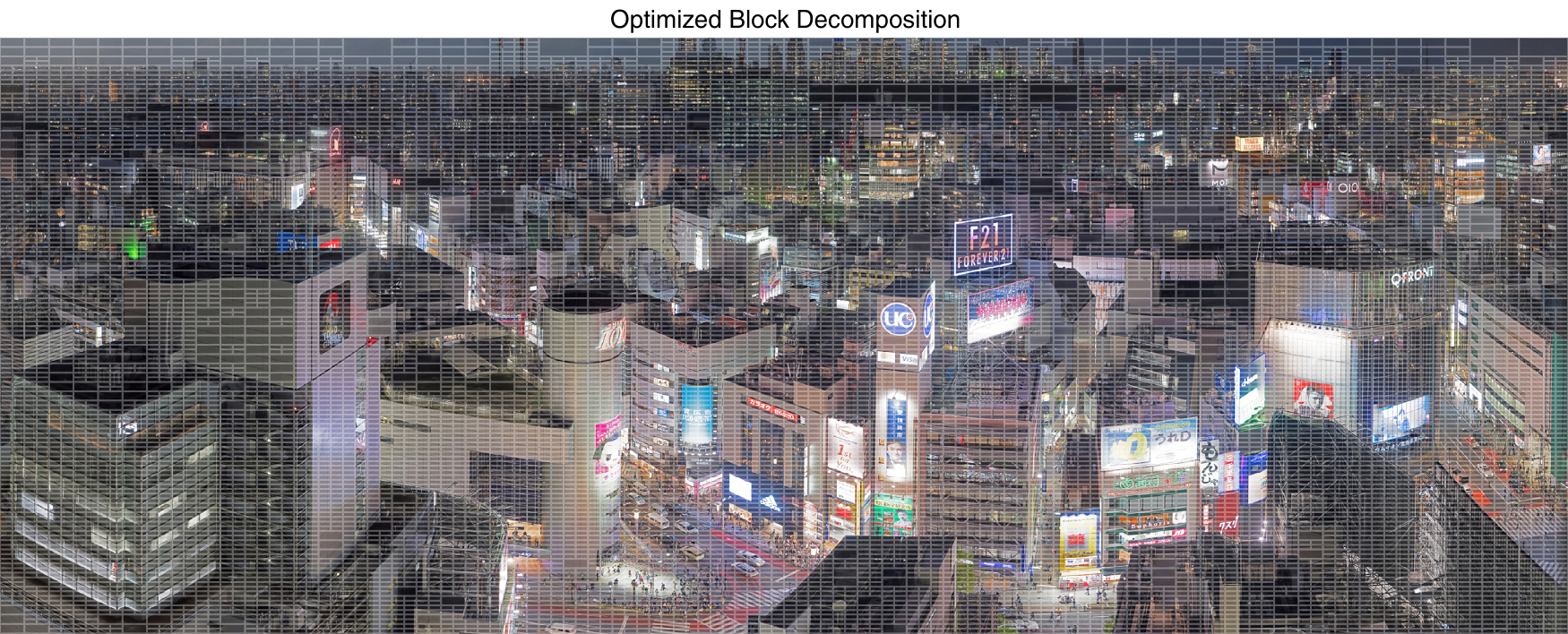}
    \caption{Optimized block decomposition for a gigapixel image. The algorithm allocates 16,384 blocks adaptively at varying scales throughout the images in order to minimize the fitting error. Blocks at various scales are shown in the overlay, with the coarsest blocks allocated to regions of flat texture and blocks at the finest scale occuring in highly detailed regions. (Photo: Trevor Dobson, \url{https://creativecommons.org/licenses/by-nc-nd/2.0/)}}
    \label{fig:supp_tokyo}
\end{figure*}

\begin{figure}[t]
    \vspace{1em}
    \includegraphics[]{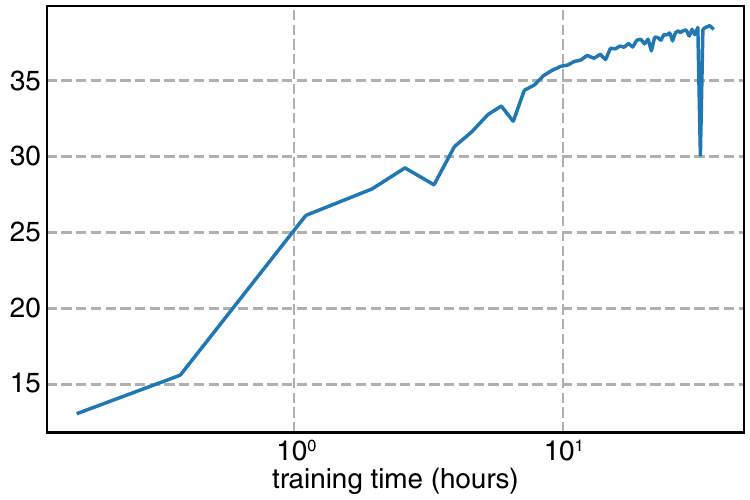}
    \caption{Plot of peak signal-to-noise ratio versus training time for fitting the gigapixel image.}
    \label{fig:supp_tokyo_psnr}
\end{figure}

\section{Supplementary Image Fitting Results}
Here we provide additional results and details on the image fitting examples shown in the main paper.
\paragraph{Pluto Image.}
In Figure~\ref{fig:supp_pluto}, we provide additional insets and error maps showing the performance of ACORN and the SIREN, and ReLU positional encoding baselines.

\paragraph{Gigapixel Image.}
We show an overlay of the block decomposition for the gigapixel image of Tokyo in Figure~\ref{fig:supp_tokyo}. 
The optimization scheme allocates larger blocks to regions of flat textures, including the background sky and sides of certain buildings. 
The finest scale blocks can be found in highly detailed regions, including text on the signage.

Finally, we also include a plot of the image fitting PSNR of training time. 
We train the network on an NVIDIA RTX 8000 GPU and find that it takes less than an hour for the representation to reach 30 dB peak signal-to-noise ratio (PSNR). 
After roughly 30 hours of training the quality improves up to roughly 39 dB PSNR. 

\begin{figure}
\includegraphics[width=0.85\columnwidth,trim=0cm 0.1cm 0cm 0cm]{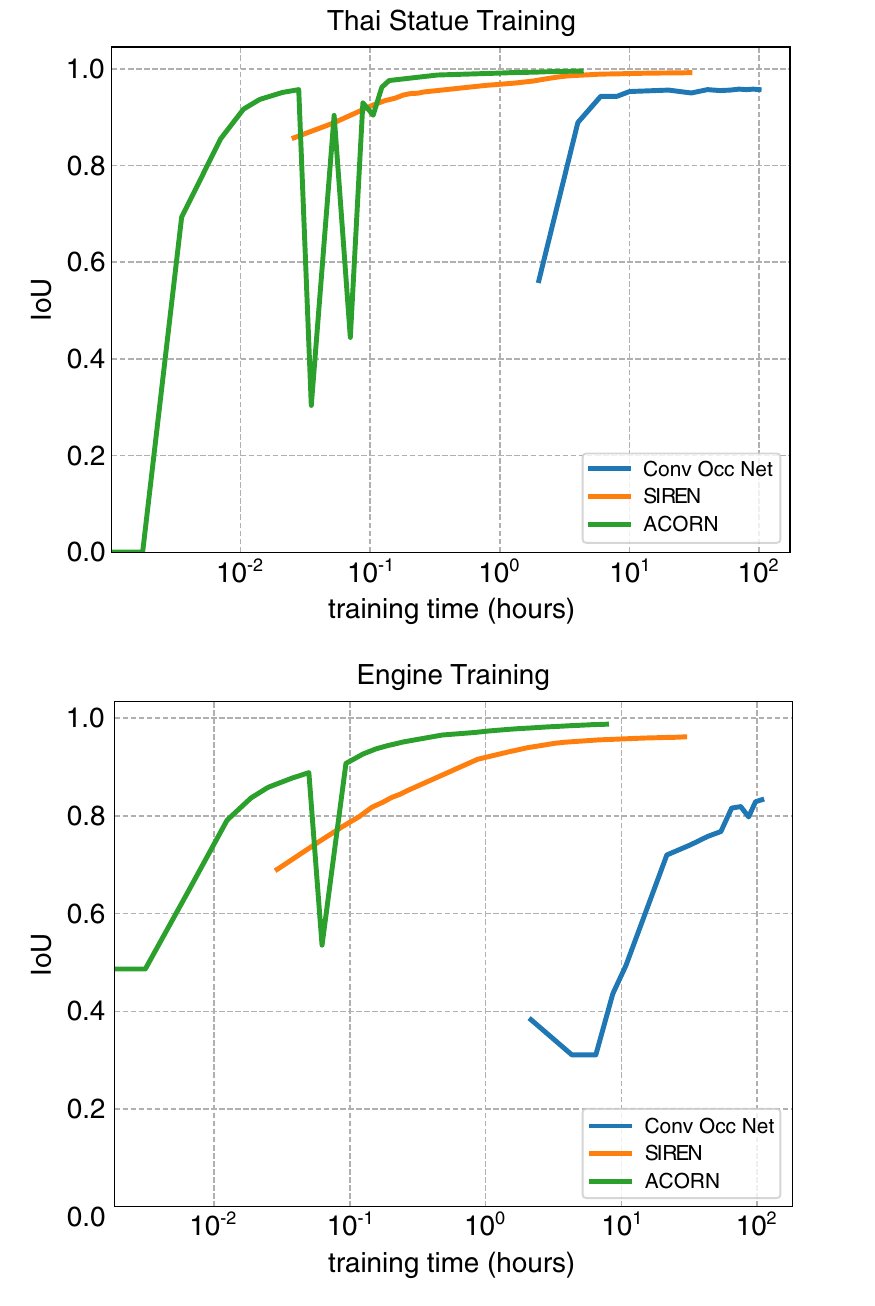}
\caption{Plot of training time versus Intersection over Union for Convolutional Occupancy Networks, SIREN, and ACORN. The proposed method trains and converges faster than the other baselines.}
\label{fig:training_iou}
\end{figure}

\begin{figure*}[t!]
    \includegraphics[trim=0cm 7cm 0cm 0cm,clip]{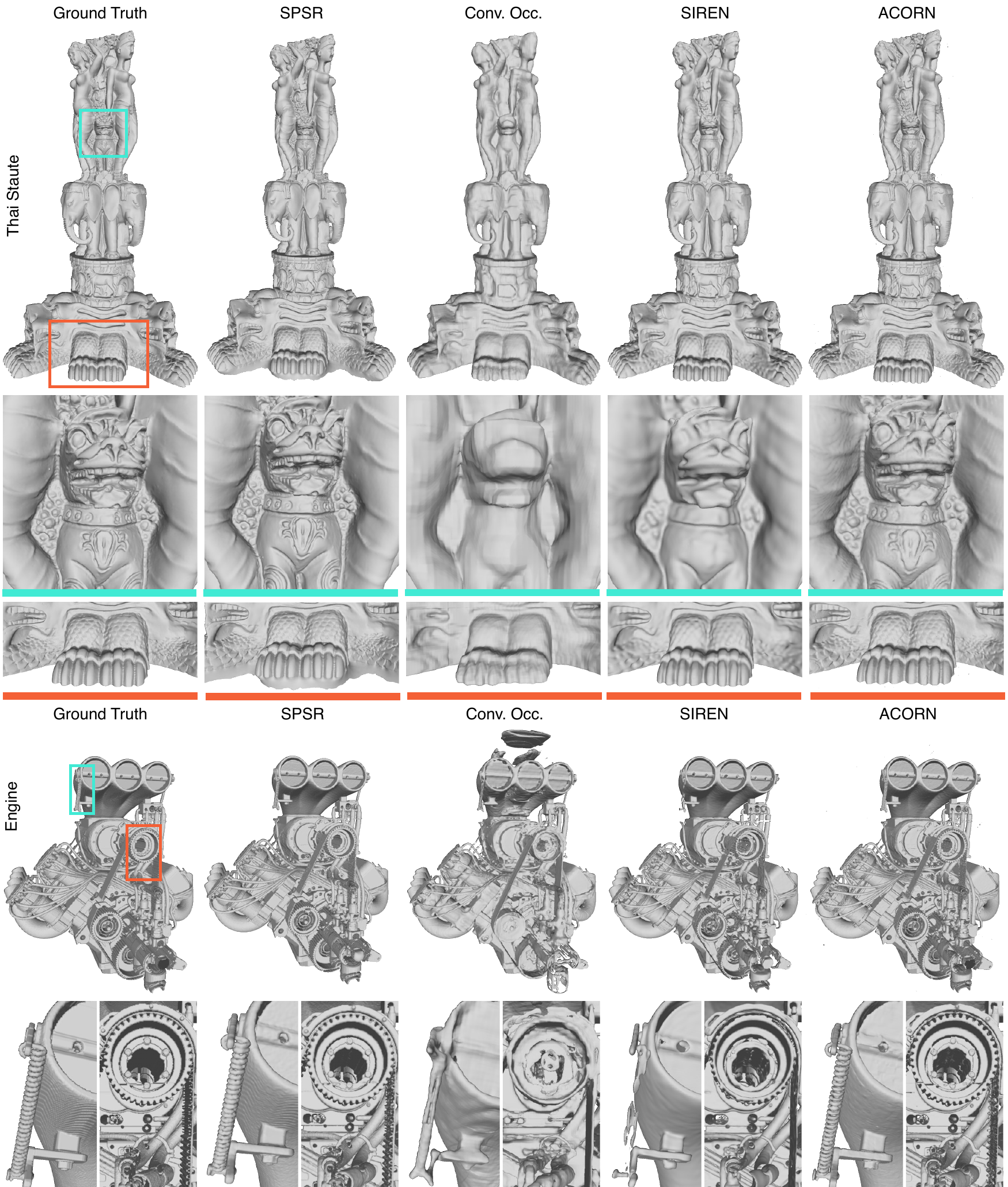}
    \caption{Large 3D scene fitting. Our method captures geometric details better than other neural methods and matches SPSR, which performs well with many uniformly sampled points, but also requires normal vectors as input. (Engine Model: ChrisKuhn, \url{https://creativecommons.org/licenses/by/2.0/})} 
	\label{fig:thai_engine}
\end{figure*}
\section{Supplementary 3D Occupancy Representation Results}
\paragraph{IoU Over Time}
Figure~\ref{fig:training_iou} plots Intersection over Union against time for ACORN and baselines, illustrating the advantage of ACORN in both performance and convergence speed. Large drops in IoU during training are due to ACORN's optimization procedure. While the reshuffling of blocks causes a short-term deterioration in representation quality, reallocating the blocks allows for a more optimal and efficient representation in the long-term.
\section{3D Scene Representation Baseline Details}
\label{sec:baselines}
In the following we provide additional implementation details for the 3D shape reconstruction baselines from the main paper. \highlight{Additional qualitative results are shown in Figure~\ref{fig:thai_engine}.}

\paragraph{Training Setup.} We first center and normalize the input mesh coordinates to fall within $[-1, 1]$ along the widest dimension (preserving the aspect ratio).
For Convolutional Occupancy Networks, we uniformly sample 5 million surface points for the input point cloud and apply Gaussian noise with standard deviation of 0.05. For the ground truth occupancy points, we sample 4.5 million surface points and apply Gaussian noise with a standard deviation of 0.01 to help the model learn better surface details.
We augment this ground truth dataset with an additional 500,000 points sampled within the bounding box of the mesh.
The ground truth occupancy is calculated for each point in the dataset and used to train the model.
For SIREN we only need the ground truth occupancy points, but we sample 20 million near surface points, and 20 million points from the bounding box of the mesh.

\paragraph{Convolutional Occupancy Networks} 
Convolutional Occupancy Networks learn a scalable implicit representation of 3D scenes by following a fully-convolutional architecture that allows for a sliding-window style of training and inference on voxel crops. Features are encoded in a discrete grid and occupancies are then predicted within the local voxel crop via a convolutional decoder.

To train the network, we sample 2048 query points inside each voxel crop, where each voxel has a size of 0.01 and each voxel crop is 25 x 25 x 25 voxels. The architecture and all other parameters are the same as the original paper's model for reconstructing large-scale 3D scenes. Finally, we train the model for 50,000 iterations. We adapt the original authors' code for our baseline implementations (\url{https://github.com/autonomousvision/convolutional_occupancy_networks}). 

\paragraph{SIREN}

Sinusoidal Representation Networks (SIRENs)~\cite{sitzmann2020siren} leverage periodic activation functions for implicit neural representations of 3D scenes. While the original implementation of SIREN for 3D objects was trained on SDF representations, we adapted the architecture to the task of occupancy prediction, with binary crossentropy loss and ground-truth occupancy values to allow for more direct comparisons. The SIREN model used in our experiments consists of four hidden layers with 2900 hidden units each, for a total of 16.8M parameters. We obtained the best performance training with a batch size of 100000 points per iteration and a learning rate of 1e-05; larger learning rates destabilized training.

\paragraph{SPSR}

Screened Poisson Surface Reconstruction (SPSR)~\cite{kazhdan2013screened} is a traditional 3D reconstruction technique which takes in an oriented point cloud and produces a reconstructed 3D mesh. Unlike our method and the other baselines, with the exception of SDF SIRENs, SPSR also requires the associated normal vectors.
To run SPSR for an input 3D mesh, we uniformly sample 5 million surface points and obtain their associated face normals. We use the same trimming parameter (--trim 6) as Peng et al.~\cite{peng2020convolutional} to fine-tune the reconstructed mesh.

\paragraph{Mesh Extraction.} We sample 20 million surface points as input and use the same voxel size and voxel crop resolution as during training. The model predicts occupancies inside each voxel crop at a pre-defined local resolution such that the combined resolution over all voxel crops is 1024 x 1024 x 1024. The mesh is then extracted from the predicted occupancies using Marching Cubes~\cite{lorensen1987marching}.

\clearpage
\bibliographystyle{ACM-Reference-Format}
\bibliography{references}